\documentclass[10pt,twocolumn,letterpaper]{article}

\usepackage{cvpr}
\usepackage{times}
\usepackage{epsfig}
\usepackage{graphicx}
\usepackage{amsmath}
\usepackage{amssymb}
\usepackage{booktabs}

\usepackage[pagebackref=true,breaklinks=true,letterpaper=true,colorlinks,bookmarks=false]{hyperref}

\cvprfinalcopy % *** Uncomment this line for the final submission

\def\cvprPaperID{****} % *** Enter the CVPR Paper ID here

\ifcvprfinal\fi

\begin{document}

%%%%%%%%% TITLE
\title{Charades-Ego: A Large-Scale Dataset of Paired Third and First Person Videos}

\newcommand{\myurl}[1]{\url{#1}}
\renewcommand{\thefootnote}{\fnsymbol{footnote}}
\author{Gunnar A. Sigurdsson $^1\footnotemark[1]$ \ \ \ \ 
Abhinav Gupta $^{1}$ \ \ \ 
Cordelia Schmid $^{2}$ \ \ \ 
Ali Farhadi $^{3}$ \ \ \ 
Karteek Alahari $^{2}$ \\
$^1$Carnegie Mellon University~~~~ \  
$^2$Inria$\footnotemark[2]$~~~~ \ \ 
$^3$Allen Institute for Artificial Intelligence \\
\myurl{allenai.org/plato/charades}\vspace{-0.5cm}
}

\maketitle
\thispagestyle{empty}

%%%%%%%%% ABSTRACT
\begin{abstract}
In~\cite{actorobserver} we introduced a dataset linking the first and third-person video understanding domains, the Charades-Ego Dataset. 
In this paper we describe the egocentric aspect of the dataset and present annotations for Charades-Ego with 68{,}536 activity instances in 68.8 hours of first and third-person video, making it one of the largest and most diverse egocentric datasets available. Charades-Ego furthermore shares activity classes, scripts, and methodology with the Charades dataset~\cite{charades}, that consist of additional 82.3 hours of third-person video with 66{,}500 activity instances. Charades-Ego has temporal annotations and textual descriptions, making it suitable for egocentric video classification, localization, captioning, and new tasks utilizing the cross-modal nature of the data. 
\end{abstract}

\footnotetext{\footnotemark[1]Work was done while Gunnar was at Inria.}
\footnotetext{\footnotemark[2]Univ.\ Grenoble Alpes, Inria, CNRS, Grenoble INP, LJK, 38000 Grenoble, France.}

%%%%%%%%% BODY TEXT
\section{Introduction}

In recent years, the field of egocentric action understanding~\cite{lee2012discovering,Li_2015_CVPR,pirsiavash2012detecting,ryoo2013first,jayaraman2015learning,Rhinehart_2017_ICCV} has exploded due to a variety of applications, including augmented and virtual reality. If we can create a link between third and first person video understanding, then we can use billions of easily available third-person videos to improve egocentric video understanding. 

In the \emph{Charades-Ego dataset} we followed the ``Hollywood in Homes'' approach~\cite{charades}, originally used to collect the Charades dataset~\cite{charades,challenge}. That is, we recruited crowd workers over the internet to record themselves in both first and third-person performing a given script of activities.

\begin{figure}
    \centering
    \includegraphics[width=1.0\linewidth]{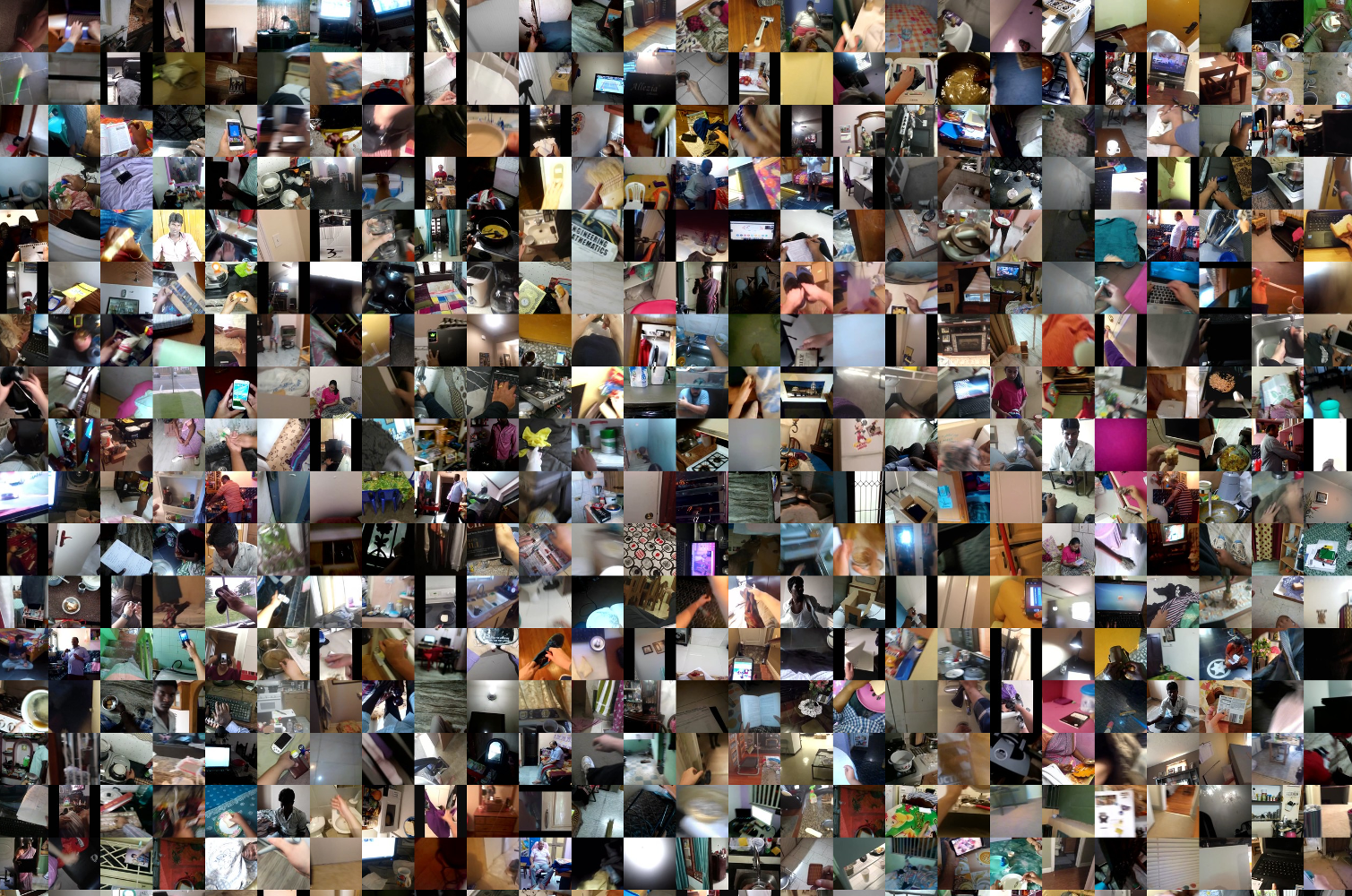}
    \caption{Example frames from the Charades-Ego dataset.}
    \label{fig:examples}
\end{figure}

Whereas~\cite{actorobserver} focused on the multi-domain aspects of the dataset. In this paper we describe the egocentric side of the dataset, and present an egocentric dataset with 68{,}536 activity instances in 68.8 hours of first and third-person video. Charades-Ego furthermore shares activity classes, scripts, and methodology with the Charades dataset~\cite{charades}, that consist of 82.3 hours of third-person video with 66{,}500 activity instances. Example egocentric frames are presented in Figure~\ref{fig:examples}, and video visualizations are available on the website.
% visual examples

\section{The Charades-Ego Dataset}
% choice of scripts
In this section we discuss the egocentric aspects of the Charades-Ego dataset, for more information about the dataset, particularly the multi-modal nature of the dataset, please refer to~\cite{actorobserver}.
We asked users on Amazon Mechanical Turk to record two videos: (1) one with them acting out the script from the third-person; and (2) another one with them acting out the same script in the same way, with a camera fixed to their forehead. This offers a compromise between a synchronized lab setting to record both views simultaneously, and scalability. In fact, our dataset is one of the largest first-person datasets available~\cite{fathi2011learning,lee2012discovering,pirsiavash2012detecting,firstthird2017cvpr}, has significantly more diversity (112 actors in many rooms), and most importantly, is the only large-scale dataset to offer pairs of first and third-person views that we can learn from.

Most of the scripts in Charades-Ego come from the Charades dataset. In addition, to ensure sufficient diversity in Charades-Ego, we collected an additional 1000 scripts using the method outlined in the Charades dataset~\cite{charades}. Therefore, $78.7\%$ of the scripts in Charades-Ego are shared with the training set of the Charades dataset. 
The users are ``given the choice to hold the camera to their foreheads, and do the activities with one hand, or create their own head mount and use two hands. We encouraged the latter option by incentivizing the users with an additional bonus for doing so. We compensated AMT workers \$1.5 for each video pair, and \$0.5 in additional bonus. This strategy worked well, with over 60\% of the submitted videos containing activities featuring both hands, courtesy of a home-made head mount holding the camera.''~\cite{actorobserver}.

% selection of activities
The activities are the same 157 activities as in the Charades dataset. In total there are 34.4 hours of third person video and another 34.4 hours of corresponding first person videos consisting of 68{,}536 activity instances, or $8.72$ activities per video on average. For comparison, the largest concurrent egocentric dataset EGTEA Gaze+, has 28 hours (10{,}325 instances)~\cite{Li_2015_CVPR}. The videos come from various rooms in 112 homes across the world.

We split the videos into 80/20 training set (6167 videos) and test set (1693 videos), ensuring that no subject occurs in both sets. The smallest number of examples per category in training and test sets is 52 and 24 examples, respectively.

\section{Video annotation}

To get a pool of activities that are likely to be present in the videos, we both copy over the activities from videos in the original Charades dataset that share the same script, as well as run the third person videos through video-level annotation following the methodology for the Charades dataset~\cite{sigurdsson2016much}.

\begin{figure}[tbhp]
    \centering
    \includegraphics[width=1.0\linewidth]{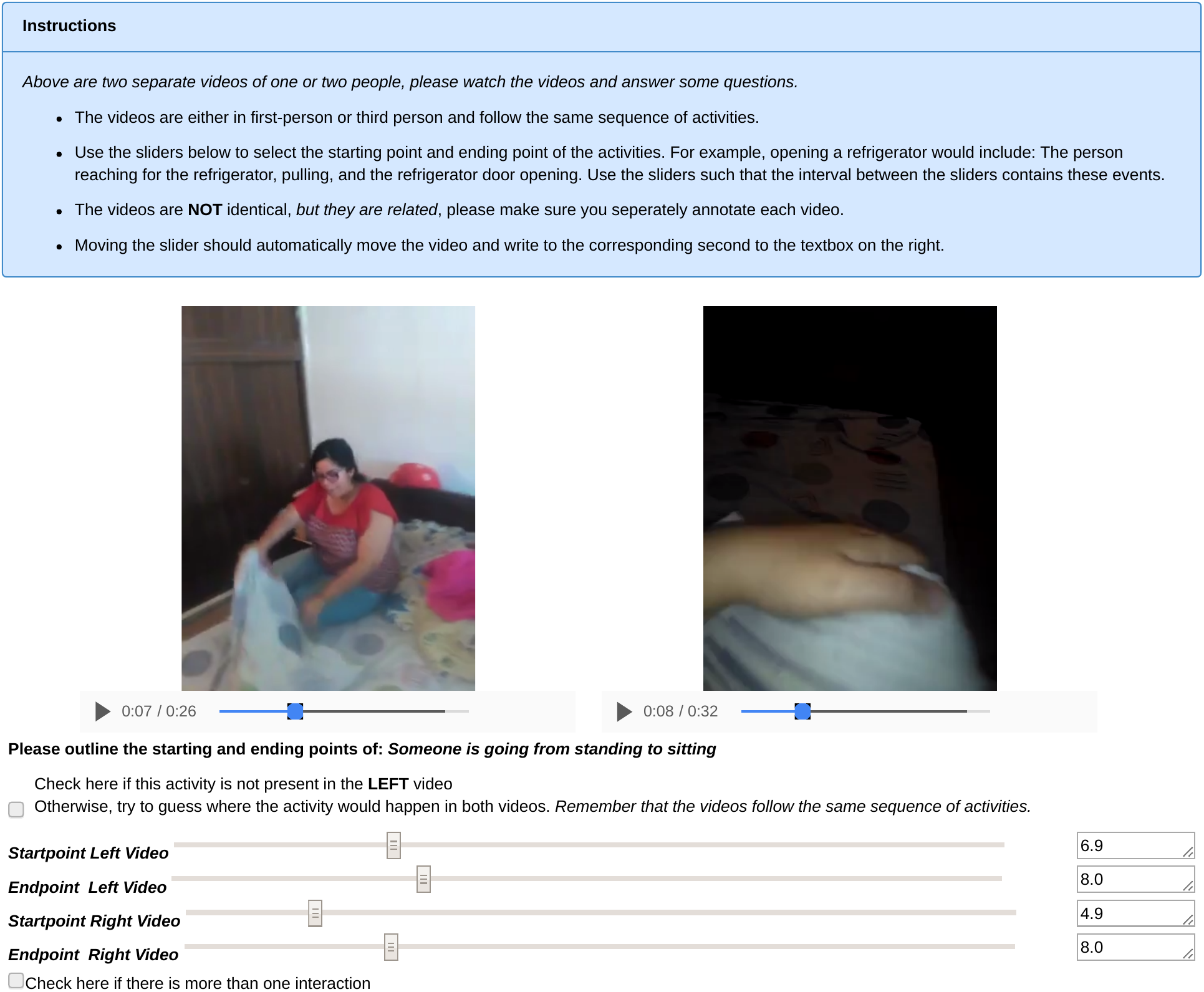}
    \caption{The interface used to collect annotations for the joint third and first person dataset. The videos are annotated simultaneously. Moving the sliders seeks in the corresponding video to help with annotation.}
    \label{fig:interface}
\end{figure}

The videos were temporally annotated by presenting both third and first person videos of from the same subject at the same time, and having the workers annotate both. The third-person video is used to obtain better annotations for the first-person video. For example, the consensus among human annotators for egocentric videos is only 4.2\% lower than for the fully-observable third-person videos (77.0\% versus 72.8\%). Each annotation task consisted of a single pair of videos and 5 activities for annotation on average. The annotation interface is presented in Figure~\ref{fig:interface}. 

\section{Baselines on Charades-Ego v1.0}

% ActorObserverNet with supervision

% egocentric baselines
To encourage research on both first and third person videos we present baselines trained on both first and third person videos and tested on either first or third person videos. These experiments are presented in Table~\ref{tbl:thirdfirst}. All models start with the same ResNet-152 models pretrained on the original Charades third-person dataset\footnote{From \myurl{github.com/gsig/charades-algorithms}.}. \emph{Third-Person Training} refers to using that original model, and \emph{First-Person Training} or \emph{First/Third Training} refers to using the labelled data in Charades-Ego for fine-tuning. We see that the different kinds of training do not improve third-person accuracy when we already have a good third-person model. However, we clearly obtain a significant improvement when using the first-person labelled data and testing on the egocentric test set.
%
% baseline training on both third and first person, testing on either first or third or both
\begin{table}[tbhp]
\centering
\caption{Comparison of training and testing on third or first person data on Charades-Ego v1.0 (video-level mean average precision)}
\label{tbl:thirdfirst}
\resizebox{\linewidth}{!}{%
\begin{tabular}{@{}lll@{}}
\toprule
                      & First-Person Test & Third-Person Test \\ \midrule
Random & 7.1 & 7.2 \\ 
Third-Person Training & 19.5              & 23.3              \\
First-Person Training & 27.1              & 17.5              \\
First/Third Training  & 28.2              & 23.2              \\ \bottomrule
\end{tabular}%
}
\end{table}

To establish an egocentric benchmark on the Charades-Ego dataset we present results of multiple baselines on v1.0 of the Charades-Ego Egocentric test set. This is presented in Table~\ref{tbl:egocentric}. Following~\cite{actorobserver}, we also present results for baselines that have not been trained with first-person labelled data, that is, zero-shot egocentric recognition. We include the baseline ResNet-152 Transfer, which uses the Charades model to predict the activities in the third person video, and then uses those labels as supervision for the first-person video.
\begin{table}[tbhp]
\centering
\caption{Comparison of egocentric baselines on the Charades-Ego v1.0 egocentric test videos (video-level mean average precision)}
\label{tbl:egocentric}
\resizebox{\linewidth}{!}{%
\begin{tabular}{@{}ll@{}}
\toprule
                      & First-Person Test        \\ \midrule
Random & 7.1 \vspace{.5em} \\
{\bf Zero-Shot} & \\
VGG-16 Charades~\cite{actorobserver}       & 15.6                        \\
ResNet-152 Charades~\cite{actorobserver}   & 19.5                     \\
ResNet-152 Transfer   & 22.2                     \\
ActorObserverNet~\cite{actorobserver} & 25.2                 \vspace{.5em} \\
{\bf Supervised on Charades-Ego} & \\
ResNet-152 Charades-Ego & 28.2                        \\ \bottomrule
\end{tabular} %
}
\end{table}

% annotated examples

\section{Discussion}

We hope this type of data is a step in bringing the fields of third-person and first-person activity recognition together. 

\paragraph{Acknowledgements}
The authors would like to thank Yin Li, Nick Rhinehart, and Kris Kitani for suggestions and advice on the dataset.

{\small
\bibliographystyle{ieee}
\bibliography{egocentric}
}

\end{document}